\documentclass[sigconf]{acmart}

\usepackage[utf8]{inputenc}
\usepackage{blindtext, graphicx}
\usepackage[utf8]{inputenc}
\usepackage[english]{babel}
\usepackage{amsmath}
\usepackage{mathtools}
\usepackage{subcaption}
\usepackage{pgffor}
\usepackage{enumerate}
\usepackage{enumitem} 
\usepackage{babel}
\usepackage{chemgreek,textgreek}
\usepackage{tikz}
\usepackage{pifont}
\newcommand{\cmark}{\ding{51}}%
\newcommand{\xmark}{\ding{55}}%
\usepackage{tikz}
\usepackage{chemgreek,textgreek}
\usepackage{tikz}
\usepackage{pifont}
\usepackage{multirow}
\usepackage[normalem]{ulem}
\useunder{\uline}{\ul}{}
\usepackage{array}
\usepackage{float}
\usepackage{graphicx,wrapfig,lipsum}
\usepackage{amsmath,hyperref}
\usepackage{multicol}
\usepackage{xcolor}

\setcopyright{acmcopyright}

\copyrightyear{2018} 
\acmYear{2018} 
\setcopyright{acmcopyright}
\acmConference[WebSci '18]{10th ACM Conference on Web Science}{May 27--30, 2018}{Amsterdam, Netherlands}
\acmBooktitle{WebSci '18: 10th ACM Conference on Web Science, May 27--30, 2018, Amsterdam, Netherlands}
\acmPrice{15.00}
\acmDOI{10.1145/3201064.3201103}
\acmISBN{978-1-4503-5563-6/18/05}

\begin{document}

\title{A Quality Type-aware Annotated Corpus and Lexicon for Harassment Research}

\author{Mohammadreza Rezvan}
\affiliation{%
  \institution{Kno.e.sis Center}
  \streetaddress{P.O. Box 1212}
  \city{Dayton}
  \state{Ohio}
  \country{USA}
  \postcode{43017-6221}
}
\email{rezvan@knoesis.org}

\author{Saeedeh Shekarpour}
\affiliation{%
  \institution{University of Dayton}
  \streetaddress{P.O. Box 1212}
  \city{Dayton}
  \state{Ohio}
  \country{USA}
}
\email{sshekarpour1@udayton.edu}

\author{Lakshika Balasuriya}
\affiliation{%
  \institution{Kno.e.sis Center}
  \streetaddress{P.O. Box 1212}
  \city{Dayton}
  \country{USA}
  \state{Ohio}
}
\email{lakshika@knoesis.org}

\author{Krishnaprasad Thirunarayan}
\affiliation{%
  \institution{Kno.e.sis Center}
  \streetaddress{P.O. Box 1212}
  \city{Dayton}
  \state{Ohio}
  \country{USA}
}
\email{tkprasad@knoesis.org}

\author{Valerie L. Shalin}
\affiliation{%
  \institution{Kno.e.sis Center}
  \streetaddress{P.O. Box 1212}
  \city{Dayton}
  \state{Ohio}
  \country{USA}
}
\email{valerie@knoesis.org}

\author{Amit Sheth}
\affiliation{%
  \institution{Kno.e.sis Center}
  \streetaddress{P.O. Box 1212}
  \city{Dayton}
  \state{Ohio}
  \country{USA}
}
\email{amit@knoesis.org}

\begin{abstract}
A quality annotated corpus is essential to research.
Despite the recent focus of the Web science community on cyberbullying research, the community lacks standard benchmarks.
This paper provides both a quality annotated corpus and an offensive words lexicon capturing different types of harassment content:
(i) sexual, (ii) racial, (iii) appearance-related, (iv) intellectual, and (v) political\footnote{Disclaimer: This paper is concerned with violent online harassment. To describe the subject at an adequate level of realism, examples of our collected tweets involve violent, threatening, vulgar and hateful speech language in the context of racial, sexual, political, appearance and intellectual harassment. 
While these examples are shared to portray the reality, the readers are alerted in advance and may wish to avoid reading this material if it could cause discomfort and disagreement.
}. 
We first crawled data from Twitter using this content-tailored offensive lexicon. As mere presence of an offensive word is not a reliable indicator of harassment, human judges  annotated tweets for the presence of harassment. 
Our corpus consists of 25,000 annotated tweets for the five types of harassment content and
 is available on the Git repository\footnote{\allowbreak{\url{https://github.com/Mrezvan94/Harassment-Corpus}}}.
\end{abstract}




\fancyhead{}  
\maketitle

\section{Introduction}


Social media is being used extensively by people from various age-groups (e.g., 80+\% usage for young adults (18-49) and 45+\% usage for old adults (50+)\footnote{Observed statistics on January 8, 2018, from Pew research at \url{http://www.pewinternet.org/fact-sheet/social-media/}.}). 
Despite the communication advantages, participants may experience insult, humiliation, bullying, and harassing comments from strangers, colleagues or anonymous users. 
One-in-five, around 18\% are affected\footnote{\scriptsize \url{http://www.pewinternet.org/2017/07/11/online-harassment-2017/}}), posing numerous challenges to
social engagement and trust, resulting in emotional
distress, privacy concerns and threats to physical safety.
All instances of harassment necessarily reflect a combination of sender intentionality and recipient experience.  
Our focus here is on the sender, whose messages are intended to harass. We study harassment\cite{cite7} in
five content areas: (i) sexual, (ii) racial, (iii) appearance-related, (iv) intellectual, and (v) political.
Below, we briefly describe each type.

\begin{itemize}
    \item \textbf{Sexual harassment} concerns sexuality and often targets females. The harasser might refer to a victim's sex organs with slang or describe sexual relations with slang.
    However, slang itself is not sufficient to indicate sexual harassment\footnote{\scriptsize\url{https://www.joshuafriedmanesq.com/sexual-harassment.html}} \footnote{\scriptsize\url{https://www.eeoc.gov/laws/types/sexual_harassment.cfm}}.

    \item \textbf{Racial harassment} targets race and ethnicity characteristics of a victim such as color, country, culture, faith, and religion\footnote{\scriptsize\url{https://www.joshuafriedmanesq.com/racial-slurs-and-racial-harassment.html}}. 

    \item \textbf{Appearance-related harassment} is related to body appearance apart from sexuality. All dimensions of appearance are candidates, for example, hair style or looks. Fat shaming \cite{cite1} and body shaming 
    are critical sub-types.

    \item \textbf{Intellectual harassment} concerns  intellectual power or the merits of individual opinion. Sub-types include level of formal education and grammar. Victims may in fact be intellectually gifted\footnote{\scriptsize\url{http://www.corrections.com/news/article/26649-ranking-bully-types-the-points-system}}.

    \item \textbf{Political harassment} relates to political views\footnote{\scriptsize\url{http://www.brighthub.com/office/career-planning/articles/89787.aspx}}, regarding issues under governmental influence such as global warming, the opiod epidemic, immigration or gun control. Typical targets are politicians and politically active individuals\footnote{\scriptsize\url{https://www.performanceicreate.com/political-discrimination-harassment/}}.
\end{itemize}

The absence of a quality, annotated corpus of online harassment impedes comparative research on harassment detection.
Our work \cite{cite7} pioneers the content-specific study of cyberbullying.  We publish here (i) our annotated content-specific lexicon and (ii) our content-specific annotated corpus validated by inter-rater reliability statistics.
This paper is organized as follows: Section \ref{sec:relatedwork} reviews the related work. Section \ref{sec:lexicon} explains the process for developing the five content-specific lexicons. In Section \ref{sec:CorpusDevelopment}, we present the strategies for collecting and annotating our corpus.  Section \ref{sec:sample} provides examples of harassing as well as non-harassing tweets. We close with concluding remarks and our future plans.

\section{Related work}
\label{sec:relatedwork}
Cyberbullying refers to the use of abusive language in social media or online interactions. While the majority of the prior research focuses on methods for detecting cyberbullying, there is no standard benchmark to evaluate and compare the performance of the existing approaches. 
The publicly available Golbeck corpus \cite{cite2} contains 25,000 unique tweets with the binary annotation labels (i.e., harassing H or non-harassing N).
There, authors use harassment hashtags such as \#whitegenocide, \#fuckniggers, \#WhitePower, and \#WhiteLivesMatter as crawling seeds. Human judges annotate the tweets using the binary labeling scheme. 

Another harassment related dataset \cite{cite8} focuses on racism and sexism. This dataset was collected during two months when the authors manually identified related hateful terms targeting groups based on aspects such as, 
ethnicity, sexual orientation, gender, and religion. 
Another corpus \cite{cite9} distinguished between cyberbullying and cyber-aggression. Collection occurred from June 2016 till August 2016 with snowball sampling.
This dataset contains 9,484 tweets from 1,303 users. Crowdsourcing workers labeled tweets according to four categories: 1) bullying, 2) aggressive, 3) spam, and 4) normal.

With respect to the methods for detecting harassment, \cite{cite3} predicts cyberbullying incidents in Instagram.
They extract features from text content, and the neighboring network along with temporal attributes to feed the predictive model. 
Another approach employed in \cite{cite4} detects harassment features using content, sentiment, and context. 
These contextual features extracted from discussion and conversation improve the accuracy of harassment detection. 
Extending the context focus, \cite{cite5} applies machine learning for detecting harassers and victims in a given cyberbullying incident. 
Their method considers social connections and infers which participants tend to bully and which participants are victimized. Their model is based on the connectivity of the users (network), the user interactions and the language of the active users.

\section{Compiling an Offensive Words Lexicon }
\label{sec:lexicon}
The identification of cyberbullying typically begins with a lexicon of potentially profane or offensive words.
We created a lexicon (compiled from online resources\footnote{\scriptsize\url{http://www.bannedwordlist.com/lists}.} \footnote{ \scriptsize\url{https://www.cs.cmu.edu/~biglou/resources}.} \footnote{\scriptsize\url{http://www.noswearing.com/dictionary}.} \footnote{\scriptsize\url{http://www.rsdb.org/races\#iranians}.} \footnote{\scriptsize  \url{http://www.macmillandictionary.com/us/thesaurus-category/american/offensive-words-for-people-according-to-nationality\\ -or-ethnicity}.}) containing offensive (i.e., profane) words covering five different types of harassment content. 
The resulting compiled lexicon includes six categories: (i) sexual, (ii) racial,   (iii) appearance-related,  (iv) intellectual, (v) political, and  (vi) a generic category that contains profane words not exclusively attributed to the five specific types of harassment. A native English speaker conducted this categorization. 
Table \ref{tab:Lexicon} represents the statistics and examples of offensive words in each category.
\begin{table}
\begin{scriptsize}
\begin{tabular}{ |p{2cm}|p{0.6cm}|p{5cm}|  }
 \hline
  \textbf{Category}& \textbf{Count} &\textbf{Example}\\
 \hline
 \textbf{Sexual}   & 453    & assfuck, ball licker, finger fucker, Anal Annie, ass blaster\\
 \hline
 \textbf{Appearance-related}&   15  & assface, dickface, fatass, fuckface, shitface   \\
 \hline
\textbf{Intellectual}  &34 & assbag, asshat, assshit, dickbrain, dumbbitch \\
\hline
\textbf{Racial}     &168 & assnigger, beaner, Bigger, mulatto, mosshead \\
\hline
 \textbf{Political} &   23  & Cockmuncher, towelhead, dickwad, propaganda, demon\\
 \hline
 \textbf{Generic}& 44  & arsehole, cockknoker, dick, fucker, sextoy \\
 \hline
\end{tabular}
\end{scriptsize}

\caption{\scriptsize Lexicon Statistics and Examples.}
\label{tab:Lexicon}
\end{table}


\section{Corpus Development and Annotation}
\label{sec:CorpusDevelopment}
We employ Twitter as the  social media data source because of its growing public footprint\footnote{\scriptsize Twitter reports 313 million monthly active users that generate over 500 million tweets per day  \scriptsize\url{https://about.twitter.com/company}.}.
Although the size of a tweet is restricted to 140 characters, once we consider a more extensive aggregation of tweets on a specific topic, mining approaches reveal valuable insights.
We utilized the first five categories of our lexicon as seed terms for collecting tweets from Twitter between December 18th, 2016 to January 10th 2017. 
Requiring the presence of at least one lexicon item, we collected 10,000 tweets for each contextual type for a total of 50,000 tweets. As shown in Table \ref{table:statistics}, nearly half of these tweets were annotated.
However, the mere presence of a lexicon item in a tweet does not assure that the tweet is harassing because the individuals might utilize these words with a different intention, e.g., in a friendly manner or as a quote.
Therefore, human judges annotated the corpus to  discriminate harassing tweets from non-harassing tweets.
Three native English speaking annotators determined whether or not a given tweet is harassing with respect to the type of harassment content and assigned one of three labels \emph{“yes”}, \emph{“no”}, and \emph{“other”}. The last label indicates that the given tweet either does not belong to the current context or cannot be decided. Finally, we can conclude 75,000 annotation work had been done totally.

\begin{table}[H]
\begin{scriptsize}
\centering
\begin{tabular}{ p{2cm}p{2cm}p{0.7cm}p{0.7cm}}
 \hline
 \textbf{Contextual Type} & \textbf{ Annotated Tweets} & \textbf{no. \cmark}  & \textbf{no. \xmark}   \\
 \hline
 Sexual    & 3855 & 230 &  3619   \\
 Racial  & 4976   &  701  &  4275\\
 Appearance-related  & 4828 & 678 & 4150  \\
 Intellectual    & 4867 & 811 & 4056  \\
 Political  & 5663   & 699 & 4964 \\ \hline
 Combined  & 24189   & 3119 & 21070 \\
 \hline
 \end{tabular}

 \caption{\scriptsize Annotation statistics of our categorized corpus. }
\label{table:statistics}
\end{scriptsize}
\end{table}

\paragraph{\textbf{Agreement Rate}} Although the annotators employed three labels, i.e., \emph{“yes”}, \emph{“no”}, and \emph{“other”}, the eventual corpus excluded all of the tweets that did not have a consensus label of \emph{“yes”} and \emph{“no”}. 
In other words, the corpus only contains tweets that receive at least two \emph{“yes”} or two \emph{“no”} labels. 
Cohen's kappa coefficient \cite{cite6} measures the quality of our annotation by category in Table \ref{table:agreementrate}.
The appearance-related context shows the highest agreement rate whereas the political and sexual contexts have the lowest indicating that 
these contents are more challenging to judge (ambiguity is higher).

\begin{table}[H]
\begin{scriptsize}
\centering
\begin{tabular}{ p{2cm}c}
 \hline
 \textbf{Content Type} & \textbf{Agreement Rate}  \\
 \hline
 Sexual &  0.70   \\
 Racial  & 0.84  \\
 Appearance-related    &  1.00      \\
 Intellectual  & 0.80     \\
 Political  & 0.69  \\
 \hline
 \end{tabular}

 \caption{\scriptsize Agreement rate.}
\label{table:agreementrate}

\end{scriptsize}
\end{table}



\paragraph{\textbf{Comparison to Golbeck Corpus}}
\label{sec:Golbeck}
The public state-of-the-art harassment-related corpus is Golbeck corpus \cite{cite2} that only provides generic annotation, i.e., (i) harassing and (ii) non-harassing. 
This corpus contains 20,428 \textbf{non-redundant} annotated tweets of which only 5,277 are labeled as harassing.
As we require a content-sensitive corpus, we created our own corpus. 
In the following, we present the principles and strategies employed in collecting, categorizing, annotating and preparing our corpus. We also categorized the Golbeck corpus according to our lexicon. It can be observed in Table \ref{tab:statisticsGolbeck} that more than 75\% of harassing tweets are racial. This statistic confirms Golbeck's observation. While this may be an accurate reflection of the base rate, our view is that different harassment content may have different consequence. An imbalanced corpus at the foundation of our research effort could result in misses of particular import to teenage mental health, concerning sexuality, appearance and intellect. 

\begin{table}[H]
\begin{scriptsize}
\centering
\begin{tabular}{ p{2cm}p{2cm}}
 \hline
 \textbf{Contextual Type} & \textbf{ \#of Tweets}   \\
 \hline
 Sexual    & 380   \\
 Racial  & 4148   \\
 Appearance-related  & 145  \\
 Intellectual    & 381  \\
 Political  & 163   \\ 
 Non Harassing  & 41  \\
 \hline
 Total & 5277
 \end{tabular}
 \end{scriptsize}
 \caption{\scriptsize Statistics of Golbeck corpus after our annotation w.r.t. contextual type.} 
\label{tab:statisticsGolbeck}
\end{table}




\section{Samples from our Corpus}
\label{sec:sample}


Below we provide some examples from our corpus, by content area.  For each content area, we first show examples annotated as harassing with respect to the content in question.  Then we show examples with similar content which are not annotated as harassing with respect to the content in question.  
\paragraph{\textbf{\textit{Sexual Harassing}}}

\begin{itemize}

{\fontfamily{qhv}\selectfont

\item @user: and you don't gotta pay none of ya bills baby ima do all that just don't fuck another nigga or ima shoot you}



{\fontfamily{qhv}\selectfont
\item to the dumbass bitch who tried opening my front door at 4 am nigga i'll kill you if i hear you again bro. i ain't a (URL)}

\end{itemize}
\paragraph{\textbf{Sexual non-harassing}}
\begin{itemize}

{\fontfamily{qhv}\selectfont
\item make up is a form of art. i do not want to be a girl with real boobs or a vagina. i may want to do drag but two very different things.
}
{\fontfamily{qhv}\selectfont
\item hot lesbian gets a pussy pounding with toy.
}
{\fontfamily{qhv}\selectfont
\item three awesome teen babes licking each other pussies in absolute lesbian sex.
}

\end{itemize}

\paragraph{\textbf{\textit{Appearance-related Harassing}}}
\begin{itemize}

{\fontfamily{qhv}\selectfont
 \item @user @user we started killing you because our backs couldn't handle the weight of your fatass anymore.
}

{\fontfamily{qhv}\selectfont
 \item  @user @user  you dat skank postn pix of my girl sab- so show ya fuckin ugly greasy mug ya getto bitch
}



\end{itemize}
\paragraph{\textbf{\textit{Appearance-related Non-harassing}}}

\begin{itemize} 

{\fontfamily{qhv}\selectfont
\item think it's funny when girls finish a tweet with babes like they're talking down to them when really they're the same level of skank.
}
{\fontfamily{qhv}\selectfont
    \item competition time\!   follow me \& amp; retweet \& amp; you can win a petite spaff\-noshing oaty camel toe blotter out of my bin.
}
{\fontfamily{qhv}\selectfont
   \item @user: @user most insulting thing a skank can do to a woman who is worth having is mock her to a woman who isn't.É
}

\end{itemize}

\paragraph{\textbf{\textit{Intellectual Harassing}}}
\begin{itemize}


{\fontfamily{qhv}\selectfont
 \item @user what a complete disgrace of human u r.real cool wish death. no surprise from a washed up fucktard really\_\_
}

{\fontfamily{qhv}\selectfont
 \item shoutout to dumb asses who go around clicking like or rt all the hot chicks posts  no matter how stupid the shit it be.
}
\end{itemize}
 \paragraph{\textbf{\textit{Intellectual non-harassing}}}
\begin{itemize}

        {\fontfamily{qhv}\selectfont
            \item maybe this isn't sadness maybe this is just being a fuckhead \_\_
}
    
            {\fontfamily{qhv}\selectfont
                \item @user oh no i'm so sorry to hear that another one of your family members is a shithead
}
    
            {\fontfamily{qhv}\selectfont
                \item @user is doing so well and his finally feeling happy   we know his been lonely and we know how tired he is ? why can't our asses be happy.
}
\end{itemize}






 \paragraph{\textbf{\textit{Racial Harassing}}}
\begin{itemize}

{\fontfamily{qhv}\selectfont
\item @user shut the fuck up chink frog nigger.
}
{\fontfamily{qhv}\selectfont
\item @user go back off private you chink.
}
{\fontfamily{qhv}\selectfont
\item @user @user @user shut up you stupid paki.
}
\end{itemize}
\paragraph{\textbf{\textit{Racial Non-harassing}}}
\begin{itemize}
        {\fontfamily{qhv}\selectfont
            \item @user do you know which exactly are the reasons for the police to release the paki? was it only cause of the 1 day period?!
}
        {\fontfamily{qhv}\selectfont
            \item rt @user: coming up on gmb  odious man-child  @user interviews racist pathological lying ass\-hat @user. 
}
        {\fontfamily{qhv}\selectfont
           \item @user 90\% of paki names are islamic hence they are not in urdu. while urdu itself is a mixture/copy of other languages even in urdu
}

\end{itemize}

\paragraph{\textbf{\textit{Political Harassing}}}
\begin{itemize}

{\fontfamily{qhv}\selectfont
\item \# thanksdonald for getting rid off that asshat who has been president for 8 yrsb.
}

{\fontfamily{qhv}\selectfont
\item @user how are u a jr high dickwad and president. a true leader doesn't taunt citizens who don't support him. pathetic. sad!
}

{\fontfamily{qhv}\selectfont
\item @user: you're passive aggressive petty fuckbag who values a murderer fascist like putting over our own president. you're oƒ.
}
\end{itemize}
\paragraph{\textbf{\textit{Political non-harassing}}}
\begin{itemize}
        {\fontfamily{qhv}\selectfont
        \item @user yep and that's how the democrats do it. you know they pretend to know what their doing but really couldn't tell their asses.

        \item  @user: liberals still continue to develop conspiracy theories in order to blame everyone else for having their asses hande ƒ.
        
        \item @user those 4 trump supporters we're bad asses to jump 20 black lives. i call bs.
        }
\end{itemize}

\section{Conclusion and Future Work}
\label{sec:conclusion}
In this paper, we discussed the creation of a quality tweet corpus related to harassment and annotated that with respect to the five types of harassment content
(i) sexual, (ii) racial, (iii) appearance-related, (iv) intellectual, and (v) political. 
This is the first corpus that takes content type into account.
Furthermore, we have also developed a lexicon of content-specific offensive words along with a generic category 
of offensive words. We are making this dataset available to encourage comparative analysis of 
harassment detection algorithms.
In future, we plan to employ this corpus for advancing our research on studying harasser and victim language.

\section{Acknowledgement}
Thilini Wijesiriwardene assisted in the preparation of the corpus. We acknowledge support from the National Science
Foundation (NSF) award CNS 1513721: Context-Aware Harassment Detection on Social Media.
Any opinions, findings, and conclusions, recommendations expressed in this material
are those of the author(s) and do not necessarily reflect the views of the NSF.

\vfill
\bibliographystyle{ACM-Reference-Format}
\bibliography{mybibliography}
\eject

\end{document}